\documentclass[letterpaper]{article} 
\usepackage[preprint]{aaai2027}  
\usepackage[hyphens]{url}  
\usepackage{graphicx} 
\usepackage{natbib}  
\usepackage{caption} 
\usepackage{algorithm}
\usepackage{algorithmic}

\newcommand{\pname}[1]{{DisMix}}

\usepackage{graphicx}
\usepackage{multirow}
\usepackage{capt-of}
\usepackage{colortbl}
\usepackage{hhline}
\usepackage{color}
\usepackage{tabularray}
\usepackage{makecell}
\usepackage{amsmath}
\usepackage{amsfonts}
\usepackage{fontawesome5}

\usepackage{newfloat}
\usepackage{listings}
\DeclareCaptionStyle{ruled}{labelfont=normalfont,labelsep=colon,strut=off} 
\floatstyle{ruled}
\newfloat{listing}{tb}{lst}{}
\floatname{listing}{Listing}

\usepackage{booktabs}

\title{\pname{}: Order-Aware Mixup for Medical Imaging via Disentangling Ordinal and Non-Ordinal Features}
\author{
    Dileepa Pitawela\textsuperscript{\rm 1} \quad
    Gustavo Carneiro\textsuperscript{\rm 2} \quad
    Hsiang-Ting Chen\textsuperscript{\rm 1}
}
\affiliations{
    \textsuperscript{\rm 1}AIML, Adelaide University, Australia \quad
    \textsuperscript{\rm 2}CVSSP, University of Surrey, UK\\
    \{dileepa.pitawela, tim.chen\}@adelaide.edu.au, g.carneiro@surrey.ac.uk
}

\begin{document}

\maketitle

\begin{abstract}
Image mixup is a widely adopted data augmentation strategy, yet it is ill-suited for ordinal classification tasks such as medical disease grading, where labels encode a progression of severity. By indiscriminately blending disease-severity cues (ordinal) with appearance-level variation (non-ordinal), standard mixup produces samples that distort the very ordinal structure that underpins clinical severity grading.
We introduce \pname{}, an order-aware mixup framework for ordinal classification. \pname{} disentangles ordinal and non-ordinal features via a dual-codebook VQ-VAE, allowing each subspace to be mixed independently: ordinal codes are interpolated to produce meaningful intermediate ranks, while non-ordinal codes are varied to introduce appearance diversity without corrupting the ordinal signal.
Across four medical imaging datasets, \pname{} 
shows the best aggregate performance among six image mixup baselines paired with six ordinal classifiers and remains effective under data scarcity and clinical grading variability. \\
Code: \url{https://github.com/dpitawela/DisMix}
\end{abstract}

\section{Introduction}
\label{sec:intro}
Image mixup augmentation is a widely used technique for training robust deep neural networks, improving generalization by generating intermediate samples that encourage smoother decision boundaries~\cite{imbalance_mixup_med}.
However, image mixup strategies~\cite{mixup,guidedmix,diffusemix,rankmix} are primarily developed for natural-image classification and are poorly suited for ordinal classification tasks such as medical imaging.
In this domain, labels commonly follow an ordinal structure, reflecting a graded progression of disease severity rather than discrete, mutually independent categories.
When conventional mixup is applied directly in this context, these methods indiscriminately blend ordinal cues (e.g., lesion severity) with non-ordinal visual factors such as imaging artifacts, background, or staining differences.
This entangled mixing corrupts the very structure that ordinal labels are intended to capture.
By ignoring the underlying features that carry the ordinal signal, the mixing can dilute or override rank-defining cues, producing samples with biologically implausible structures that deviate from the intended ordinal progression.

\begin{figure*}[t]
\centering
  \includegraphics[width=0.9\textwidth]{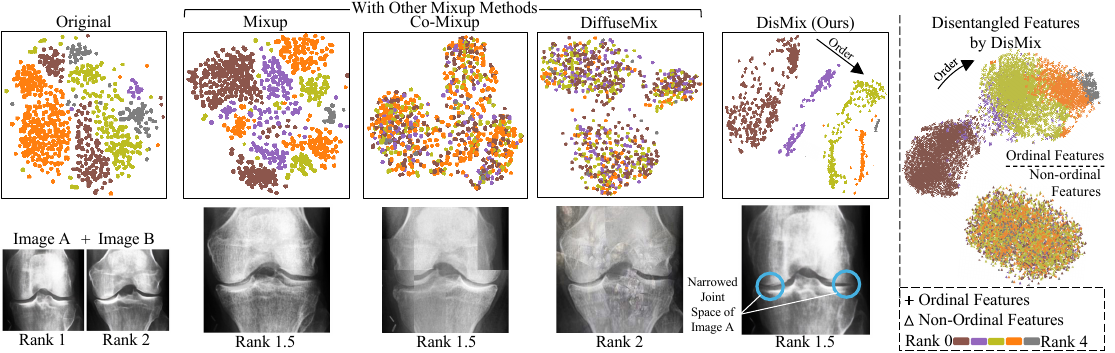}
  \captionof{figure}{
    \textbf{Main panel (left):} Original feature space of the Knee Osteoarthritis (KOA) dataset~\cite{ds_koa}, features and images from Mixup~\cite{mixup}, Co-Mixup~\cite{comix}, DiffuseMix~\cite{diffusemix}, and our \pname{}.
    \textbf{Top row:} Existing mixup entangles features and collapses the structure of data that ordinal labels are meant to capture. \pname{} preserves clusters of original data and enforces ordinal structure.
    \textbf{Bottom row:} Naive mixing produces distorted images or anatomically inconsistent structures (e.g., disoriented knee joint gaps), whereas mixing using disentangled latents, \pname{} produces plausible, order consistent samples.
    \textbf{Rightmost panel:} \pname{} disentangles ordinal and non-ordinal factors of data into separate latent spaces, learning an ordered manifold for former and a rank-invariant cluster for latter.
    (t-SNEs were produced by passing mixup images through an ILSVRC pretrained ResNet-50~\cite{tsne}).
  }\label{fig:teaser}
\end{figure*}

Fig.~\ref{fig:teaser} shows this
failure mode.
In the original Knee Osteoarthritis (KOA) dataset~\cite{ds_koa}, samples with the same rank organize into distinct clusters.
After naive mixing, this structure deteriorates: samples from different ranks become entangled, intermediate points drift away from the ordinal trajectory, and the overall feature distribution loses its ordered structure.
Rather than enriching the data distribution, these mixed samples distort it, producing anatomically inconsistent samples (e.g., disoriented knee joint gap) that may hinder the learning of robust ordinal classifiers.
In contrast, mixing using disentangled latents produces biologically plausible and order consistent samples with \pname{}.

We address the fundamental incompatibility between conventional mixup and ordinal learning,
recognizing that not all feature mixing is consistent with the underlying rank progression and that order-aware mixing is therefore necessary.
To this end, \pname{} trains a dual-codebook VQ-VAE that factorizes each image into an ordinal latent capturing rank-defining structure and a non-ordinal latent encoding rank-independent variations. The separation is enforced using soft ordinal supervision on the ordinal branch and an adversarial rank-removal objective on the non-ordinal branch.
Building on this factorization, we propose a set of order-aware mixup policies (Tab.~\ref{table:mix_policies}) that operate separately on the two subspaces, generating order-consistent augmented samples without corrupting the underlying ordinal signal.
To summarize,
\begin{itemize}
\item We introduce DisMix, an order-aware mixup method that disentangles ordinal and non-ordinal features into distinct latent subspaces, enabling synthesis of order-consistent augmented samples.
\item We propose a set of order-aware mixup policies that operate in the disentangled latent space, allowing controlled mixing during downstream ordinal classifier training.
\item We comprehensively evaluate DisMix on four medical imaging datasets, against six mixup baselines using six ordinal classifiers.
\end{itemize}

Across 24 mixup-classifier combinations, \pname{} achieves the best accuracy and MAE in 20 and 15 settings, respectively.
Compared to the strongest mixup baseline, \pname{} delivers statistically significant aggregate accuracy gains (one-sided Wilcoxon signed-rank test, $p=0.0075$) and reduces MAE by 4--6\%, while remaining robust under data scarcity and grading variability.

\section{Related Work}

\subsubsection{Ordinal Classification.}
Ordinal classification preserve class order using specialized objectives, as standard multi-class losses ignore the relationships between adjacent ranks. 
ORCNN decomposes prediction into cumulative binary thresholds~\cite{orcnn}; CNNPOR and MWR enforce pairwise and local-to-global ordinal constraints~\cite{cnnpor,mwr}; POE models boundary uncertainty through probabilistic embeddings~\cite{poe}; Ord2Seq recasts ranking as a sequence prediction~\cite{ord2seq}; and GOL learns geometry-aware ordinal representations~\cite{gol}.
To improve robustness to grading variability, SORD adopts soft labels~\cite{sord}, RNC aligns embeddings with rank order via contrastive learning~\cite{rnc}, and CLOC learns adaptive margins between neighboring ranks~\cite{cloc}.
While robust to mild grading variability, these methods do not explicitly address the grader disagreements common in medical imaging, which typically occur between adjacent ranks~\cite{cloc}, especially when ordinal cues are entangled with appearance variations.
Conventional image mixup worsens this by indiscriminately blending ordinal and non-ordinal features.
We instead disentangle these factors and apply controlled mixup: label-preserving by mixing non-ordinal factors and label-interpolating by mixing ordinal ones, improving supervision under grading variability.

\subsubsection{Disentangled Representation Learning.}
Disentangled representation learning aims to separate data into latent factors that capture distinct sources of variation.
VQ-VAEs introduce discrete codebooks promoting organized latent representations \cite{vqvae}, with conditional VQ-VAE, FactorQVAE, and QLAE extending this paradigm through conditional guidance, total-correlation regularization, and improved semantic organization \cite{cond_vqvae,factor_qvae,qlae,tripod}.
Adversarial methods disentangle factors using mutual information, attribute supervision, or pretrained generators \cite{infogan,factor_gan,dis_rep,dis_survey}.
Architectural designs further promote factorized representations \cite{style_gan,fader_net,factor_vae}, enabling applications such as anatomy/pathology separation, pose- and age-invariant face recognition, content-style decomposition, and ordinal content preservation \cite{dis_norm_ab,dr_gan,dis_face,bicycle_gan,munit,dis_aug}.
Diffusion models explore feature factorization \cite{dis_diffusion1,dis_diffusion2}, but aim high-quality generation at substantially higher computational cost.
In contrast, we use relatively light-weight VQ-VAE and disentangle ordinal and non-ordinal features from class labels alone, without attribute supervision, and perform order-aware latent mixup.

\subsubsection{Image Mixing Augmentation.}
Image mixing augments data by combining samples and labels; however, existing methods are largely semantics-agnostic on which factors are being mixed.
Vanilla Mixup linearly blends pixels and labels \cite{mixup}; CutMix pastes image patches and mixes labels by area \cite{cutmix}; PuzzleMix, Co-Mixup, and GuidedMixup preserve salient regions during mixing \cite{puzzlemix,comix,guidedmix}; and Manifold Mixup performs interpolation in latent space \cite{manifold_mix}.
RankMixup uses rank-informed labels \cite{rankmix}; SUMix learns uncertainty-aware soft labels \cite{sumix}; MOM smooths ordinal targets for manifold mixup \cite{manifold_ordmix}; and SGD-Mix and DiffuseMix propose label-preserving augmentation \cite{sgd_mix,diffusemix}, but none explicitly separate ordinal and non-ordinal features for mixing.
In ordinal settings, labels reflect ordinal features, while non-ordinal variations are expected to be irrelevant.
Prior methods mix entangled latents, either corrupting labels or under-utilizing augmentation by ignoring appearance cues.
Our method instead disentangles ordinal and non-ordinal features, enabling controlled mixing in separate latent subspaces for improved label fidelity and semantic control.

\subsubsection{Grading Variability.}
Medical datasets frequently exhibit grading variability across assessments due to inherent ambiguity in clinical interpretation \cite{carneiro2024machine,ji2021learning}.
Existing approaches address this through majority voting, probabilistic consensus models \cite{dawid1979maximum,sinha2018fast,goh_crowdlab_2023}, or by jointly estimating annotator reliability during training \cite{cao_learning_2023,herde2023multi,herde2024annot,l2cu}.
However, these methods primarily estimate consensus, have seen limited study in ordinal settings \cite{ordinal_mrl1,ordinal_mrl2}, and have not been explored for order-preserving latent mixup.
The closest related work, CLOC \cite{cloc}, learns ordinal representations under grading variability but assumes a single assessment.
In contrast, our method focuses on preserving ordinal geometry through order-aware latent mixup under grading variability.
\section{Preliminaries}
\subsubsection{Ordinal classification.}
Ordinal classification is a special case of multi-class classification where classes follow a natural ordering denoted by $r_0 \prec r_1 \prec ... \prec r_{C{-}1}$, where $C$ is the number of classes, and $\prec$ indicates the ranking relation.
Ordinal classification assumes that images are composed of ordinal features, which determine the rank, and rank-invariant non-ordinal features \cite{dis_aug}.
Beyond accuracy, ordinal consistency is measured by mean absolute error (MAE) accounting for the magnitude of rank errors. 
Throughout this paper, the terms class, label, grade, and rank are used interchangeably, while grading variability refers to disagreements among annotators.

\subsubsection{Problem Setup.} Let the training dataset be 
$\mathcal{D}{=}\{ (x_i,\{y_{i,a}\}_{a \in \mathcal{A}}) \}_{i=1}^{N}$,
where each sample $x_i {\in} \mathcal{X}$ has a set of ordinal labels
$y_{i,a} {\in} \mathcal{Y} {=} \{0,1,...,C{-}1\}$ that reflect grading variability among annotators $\mathcal{A}$.

\subsubsection{Ordinal Soft Labels.}
Standard soft labels capture annotation uncertainty but ignore ordinal relationships between classes.
Ordinal soft labels capture both uncertainty and ordinal relation.
For each $x_i$ with scalar ranks $\{y_{i,a}\}_{a \in \mathcal{A}}$, each rank is first transformed into a progressive binary vector.
These vectors are then averaged across annotators to compute the ordinal soft target $\ddot{y}_i$ with
$
\ddot{y}_i^{(k)} = \tfrac{1}{|\mathcal{A}|}\sum_{a \in \mathcal{A}} \mathbb{I}(k {<} y_{i,a}),
$
for $k \in \{0, ...,  C{-}2 \}$, where $\ddot{y}_i \in \ddot{\mathcal{Y}} \subset \mathbb{R}^{C-1}$ and $\mathbb{I}(\cdot)$ is an indicator function.
Examples are provided in App.~\ref{app:soft_ord_label}.

Ordinal supervision is provided via the Soft Ordinal Regression (SOR) loss~\cite{sord}, computed as binary cross-entropy over ordinal soft labels, which penalizes prediction errors in proportion to their ordinal distance from the target during \pname{} generator training.

\section{Methodology}
Conventional mixing of ordinal images disrupts the underlying ordinal structure by indiscriminately blending rank-defining ordinal and rank-independent non-ordinal features;
\pname{} instead learns separate latent subspaces for each, enabling independent interpolation of ordinal features to generate intermediate ranks and non-ordinal features to introduce appearance diversity without disrupting ordinal semantics.

\subsection{Disentangling Ordinal and Non‑Ordinal Factors}
\label{sec:disentangling}
\begin{figure}
    \centering
    \includegraphics[width=0.85\columnwidth]{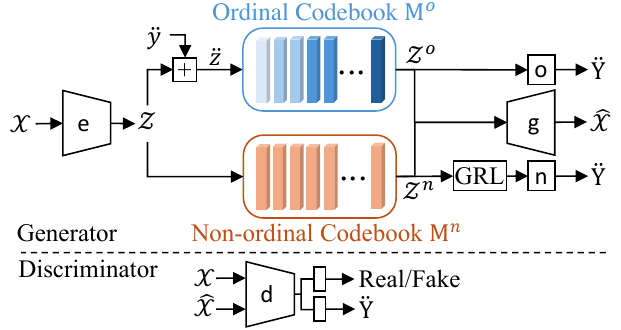}
    \captionof{figure}{
    Overview of \pname{} architecture, featuring a dual-codebook VQ-VAE generator for modeling ordinal and non-ordinal feature spaces, and a multi-task discriminator for enforcing ordinal consistency and perceptual quality.
    }
    \label{fig:gan}
\end{figure}

\pname{} employs a dual-codebook VQ-VAE (Fig.~\ref{fig:gan}) to disentangle ordinal and non-ordinal features.
Formally, the encoder $\mathsf{e}:\mathcal{X} \to \mathcal{Z}$ maps an input image $x$ to a latent feature map $z$, which is then factorized into an ordinal component $z^{\text{o}} \in \mathcal{Z}^{\text{o}}$ and a non-ordinal component $z^{\text{n}} \in \mathcal{Z}^{\text{n}}$ via the ordinal and non-ordinal branches.
Each branch maintains a dedicated discrete codebook to facilitate a compact representation.
The decoder $\mathsf{g}:\mathcal{Z}^{\text{o}} \times \mathcal{Z}^{\text{n}} \to \mathcal{X}$ reconstructs the image from the quantized codes $\hat{x} = \mathsf{g}(z^{\text{o}}, z^{\text{n}})$, enabling the model to learn meaningful decompositions while preserving visual fidelity.
The framework is optimized jointly with an adversarial discriminator $\mathsf{d}:\mathcal{X} \to \{0,1\} \times \ddot{\mathcal{Y}}$, which ensures perceptual realism and ordinal consistency.

\subsubsection{Ordinal Quantizer.}
The ordinal branch maintains a learnable codebook $\mathbf{M}^{o} \in \mathbb{R}^{|\mathcal{Z}| \times D^{o}}$, with $D^{o}$ vectors of dimension $|\mathcal{Z}|$, to capture ordinal features.
To bias towards ordinal semantics, the encoder features are conditioned on the ordinal soft label $\ddot{y}$.
Specifically, the label is projected and added to the encoder output, followed by sigmoid gating to emphasize rank-relevant channels:
$\ddot{z} = (z + h_1(\ddot{y})) \odot h_2(h_1(\ddot{y})),$
where $h_1:\ddot{\mathcal{Y}}\rightarrow\mathcal{Z}$, $h_2:\mathcal{Z}\rightarrow[0,1]^{|\mathcal{Z}|}$, and $\odot$ denotes elementwise multiplication.

Instead of the standard distance-based lookup, we adopt attention-based quantization \cite{softvq}, allowing the conditioned features to attend selectively to codebook entries by treating $\ddot{z}$ as \emph{queries} and the codebook entries as \emph{keys} and \emph{values}.
To reduce over-smoothing of ordinal features, we replace soft attention with hard attention:
$z^{o} =  \mathsf{one\_hot} \left(\text{softmax}(\ddot{z}^{\top} \mathbf{M}^{o}) \right) \left(\mathbf{M}^{o}\right)^{\top},$
where $\mathsf{one\_hot}(\cdot)$ selects the highest-attention codebook entry.
During training, hard assignments are used in the forward pass, with gradients propagated through the soft attention weights using the straight-through estimator.

To encourage the codebook to learn order-discriminative features, we attach a linear classifier 
$\mathsf{o}:\mathcal{Z}^o \to \ddot{\mathcal{Y}}$
that predicts rank from $z^{o}$.
This auxiliary supervision is imposed via the SOR loss $\ell_{\text{C}}^{\text{o}}(z^{o},\ddot{y}) = \text{SOR}(\mathsf{o}(z^{o}), \ddot{y})$
to align quantized embeddings with the ordinal structure of data.

\subsubsection{Non-Ordinal Quantizer.}
\label{sec:nonord_quant}

The non-ordinal branch maintains a learnable codebook $\mathbf{M}^{n} \in \mathbb{R}^{|\mathcal{Z}| \times D^{n}}$ 
paired with an adversarial classifier to capture order-invariant features.
Given the encoder embedding $z$, the nearest codebook entry is selected to obtain the quantized representation $z^{n}$.

To suppress ordinal information from this branch, we attach a linear classifier $\mathsf{n}:\mathcal{Z}^{n} \rightarrow \ddot{\mathcal{Y}}$ through a Gradient Reversal Layer (GRL) \cite{grl} (Fig.~\ref{fig:gan}).
The classifier attempts to infer the ordinal label from $z^{n}$, while the GRL inverts feedback during backpropagation, forcing the quantizer to suppress order-sensitive signals.
The classifier is supervised by SOR, with
$\ell_{\text{C}}^{\text{n}}(z^{n},\ddot{y})=\mathrm{SOR}(\mathsf{n}(\mathrm{GRL}(z^{n})),\ddot{y})$, 
thereby adversarially 
discouraging the non-ordinal codebook from encoding rank information.

\textit{Quantizer Losses.}
The ordinal and non-ordinal codebooks are optimized with standard vector-quantization losses $\ell_{\text{Q}}^{\text{o}}(z_1,z_2)$ and $\ell_{\text{Q}}^{\text{n}}(z_1,z_2)$~\cite{vqvae}, both following
$\|\text{sg}[z_1]-z_2\|_2^2 + \|z_1-\text{sg}[z_2]\|_2^2$, 
where $\operatorname{sg}[\cdot]$ denotes the stop-gradient.
The ordinal loss uses $(z_1,z_2)=(\ddot{z},z^{o})$, while the non-ordinal loss uses $(z_1,z_2)=(z,z^{n})$.
To discourage correlation and promote disentanglement, we additionally minimize the cosine similarity between the two latent representations,
$\ell_{\text{sim}} {=} \eta(z^{\text{o}}, z^{\text{n}})$

\subsubsection{Decoder and Discriminator.}
\label{sec:discriminator}
The decoder $\mathsf{g}:\mathcal{Z}^o \times \mathcal{Z}^n \to \mathcal{X}$ reconstructs the image from the concatenated quantized representations as 
$\hat{x} {=} \mathsf{g}(z^{o}, z^{n})$ and
is optimized to minimize pixel reconstruction error and perceptual distance with $\ell_{\text{rec}}(x,\hat{x}) {=} \|x{-}\hat{x}\|_2^2 + \text{LPIPS}(x,\hat{x})$ \cite{lpips}.

The discriminator $\mathsf{d}:\mathcal{X} \to \{0,1\} \times \ddot{\mathcal{Y}}$ adopts a PatchGAN architecture~\cite{patchgan} with shared convolutional layers and two task-specific heads.
The real/fake head $\mathsf{d}_{\text{rf}}$ is optimized with $L_2$ objective,
$\ell_{\mathsf{d}_{\text{rf}}}(x,\hat{x}) {=} \big\|\mathsf{d}_{\text{rf}}(x)-1\big\|_2^2 + \big\|\mathsf{d}_{\text{rf}}(\hat{x})\big\|_2^2$
while the classification head $\mathsf{d}_{\text{clf}}$ is supervised with SOR,
$\ell_{\mathsf{d}_{\text{clf}}}(x,\ddot{y}) {=} {\text{SOR}}\!\left(\mathsf{d}_{\text{clf}}(x),\,\ddot{y}\right)$.
Together, these objectives encourage perceptually realistic reconstructions that preserve ordinal semantics.

\subsection{\pname{} Training}

The generator, comprising encoder $\mathsf{e}(.)$, codebooks $\mathbf{M}^{o},\mathbf{M}^{n}$, and decoder $\mathsf{g}(.)$, is trained adversarially against the discriminator $\mathsf{d}(.)$ via:
\begin{align}
\min_{\mathsf{e},\mathbf{M}^{o},\mathbf{M}^{n}, \mathsf{g}}
\max_{\mathsf{d}}\;
\mathbb{E}_{x}\!\big[
\ell_{(\mathsf{e},\mathbf{M}^{o},\mathbf{M}^{n},\mathsf{g})}
-
\ell_{\mathsf{d}}
\big]
\label{eq:overall_optim}
\end{align}
The generator loss $\ell_{(\mathsf{e},\mathbf{M}^{o},\mathbf{M}^{n},\mathsf{g})}$ combines quantization ($\ell_{\text{Q}}^{\text{o}}, \ell_{\text{Q}}^{\text{n}}$), classification ($\ell_{\text{C}}^{\text{o}}, \ell_{\text{C}}^{\text{n}}$), reconstruction ($\ell_{\text{rec}}$), and similarity ($\ell_{\text{sim}}$) terms, while
the discriminator loss $\ell_{\mathsf{d}}$ combines realism ($\ell_{\mathsf{d}_{\text{rf}}}$) and ordinal classification ($\ell_{\mathsf{d}_{\text{clf}}}$) objectives.
All terms are weighted by learnable coefficients that are jointly optimized with the model under a unit-sum constraint, eliminating the need for manual hyperparameter tuning.

\subsection{Order Aware Mixup with Disentangled Latents}
\label{sec:aug_policies}
\begin{table}[tb]
\centering
\resizebox{\linewidth}{!}{%
\begin{tabular}{c|l|c} \hline
\multicolumn{1}{c|}{Policy~}  & \multicolumn{1}{c|}{Latent Interpolation} & \multicolumn{1}{l}{Label Assignment}  \\ \hline\hline
\multirow{2}{*}{\begin{tabular}[c]{@{}c@{}}Ordinal\\Mix\end{tabular}} & $\tilde{z}^o = \lambda z^{o}_i + (1{-}\lambda) z^{o}_j,$ & \multirow{2}{*}{$\tilde{y} = \lambda \bar{y}_i + (1{-}\lambda)\bar{y}_j$}                \\
                             & $\tilde{z}^n\!\in\!\{z^{n}_i,z^{n}_j\}$                 &                                       \\ \hline

\multirow{2}{*}{\begin{tabular}[c]{@{}c@{}}Non-Ord-\\inal Mix\end{tabular}} & $\tilde{z}^o\!\in\!\{z^{o}_i,z^{o}_j\},$                  & $\tilde{y}$ = $\bar{y}$ of sample                \\
& $\tilde{z}^n = \lambda z^{n}_i + (1{-}\lambda) z^{n}_j$                  &  providing $\tilde{z}^o$                                    \\ \hline

\multirow{2}{*}{\begin{tabular}[c]{@{}c@{}}Order\\Swap\end{tabular}} & $(\tilde{z}^o,\tilde{z}^n) 
\!\in\!\{(z^{o}_j,z^{n}_i),$                 & $\tilde{y}$ = $\bar{y}$ of sample \\
                             & ~ ~ ~ ~ ~ ~ ~ ~ ~ ~$(z^{o}_i,z^{n}_j)\}$                  & providing $\tilde{z}^o$                                     \\ \hline

\multirow{2}{*}{\begin{tabular}[c]{@{}c@{}}Generate\\\& Mix\end{tabular}} & $\tilde{z}^o = \lambda z^{o,\text{anc}} + (1{-}\lambda) z^{o,x'},$                  & $\tilde{y} = \lambda \bar{y}^{\text{anc}} + $             \\
                             & $\tilde{z}^n \in \{z^{n,\text{anc}}, z^{n,x'}\}$                  &      $(1{-}\lambda) y'$                                 \\ \hline


\end{tabular}
}
\captionof{table}{Latent mixing and labels assignment in mixup policies. The mixed sample is obtained by decoding $\tilde{z}^o$ and $\tilde{z}^n$.}
\label{table:mix_policies}
\end{table}
Once trained, \pname{} is frozen and used to generate mixed samples during downstream ordinal classifier training.
Given two images $x_i, x_j$ from adjacent ranks $\bar{y}_i, \bar{y}_j$ determined by majority grade (i.e., $\bar{y}_i{=}\text{majority}(\{y_{i,a}\}_{a \in \mathcal{A}})$ and similarly for $\bar{y}_j$),
with corresponding ordinal and non-ordinal codes $(z_i^{o},z_i^{n})$ and $(z_j^{o},z_j^{n})$ extracted from quantizers, and a mixing ratio $\lambda\in(0,1)$, Tab.~\ref{table:mix_policies} defines four mixing policies.

\textit{\underline{Ordinal Mix}} interpolates ordinal factors while keeping non-ordinal features fixed.
Formally,
$\tilde{z}^{o} {=} \lambda z^{o}_i {+} (1{-}\lambda) z^{o}_j, \quad$
$\tilde{z}^n {\in} \{z^{n}_i, z^{n}_j\}$.
Then we obtain the augmented sample by,
$\hat{x} {=} \mathsf{g}(\tilde{z}^{o}, \tilde{z}^{n}),$
and its label $\tilde{y} {=} \lambda \bar{y}_i {+} (1{-}\lambda) \bar{y}_j$.
\textit{\underline{Non-Ordinal Mix}} instead interpolates order-invariant features and 
takes the majority grade of the sample that provides the ordinal code.
\textit{\underline{Order Swap}} swaps the ordinal and non-ordinal features between samples.
$
(\tilde{z}^{\text{o}}, \tilde{z}^{\text{n}}) \in
\{(z^{o}_j, z^{n}_i), (z^{o}_i, z^{n}_j)\}, \quad
\hat{x} = \mathsf{g}(\tilde{z}^{\text{o}}, \tilde{z}^{\text{n}}),
$
and $\tilde{y}$ is the rank associated with the chosen $\tilde{z}^{\text{o}}$.

\textit{\underline{Generate \& Mix}} generates an adjacent-rank variant of an anchor and applies Ordinal Mix.
We first select either $x_i$ or $x_j$ as an anchor and obtain an adjacent rank from its majority grade.
By conditioning the anchor's ordinal code on the adjacent rank, we obtain $z^{o,\text{cnd}}$ from the ordinal quantizer.
The non-ordinal code $z^{n,anc}$ of the anchor is extracted via the non-ordinal quantizer to produce
$x' {=} \mathsf{g}\!\left(z^{o,\text{cnd}},\, z^{n,\text{anc}}\right),$
with $y'$ corresponding to the conditioned rank.
Ordinal Mix is then applied between the anchor and $x'$.

Non-Ordinal Mix and Order Swap are label-preserving policies, retaining the rank of the source sample providing the ordinal code.
Ordinal Mix and Generate \& Mix are label-interpolating policies, assigning soft labels between the two source ranks.
\begin{table*}[tb]
\centering
\arrayrulecolor{black}
\resizebox{\linewidth}{!}{%
\begin{tabular}{r|cccccc|ccccccc} \hline
~                     & \multicolumn{6}{c|}{Accuracy $\uparrow$}    & \multicolumn{6}{c}{Mean Absolute Error (MAE) $\downarrow$}                                                                                                                          \\ \hline
~                     & POE                   & GOL                   & MWR                   & RNC                   & ORD2SEQ               & CLOC                  & POE                  & GOL                  & MWR                  & RNC                  & ORD2SEQ              & CLOC                  \\ \hline\hline
\multicolumn{1}{r}{~} & \multicolumn{12}{c}{~ ~ ~ ~ ~ ~ ~ ~IDRID}                                                                                                                                                                                                                                                                                             \\ \hline
Mixup                 & 57.93 ± 4.05          & 55.34 ± 1.68          & 39.16 ± 1.13          & 54.04 ± 3.12          & 63.43 ± 0.56          & 64.08 ± 0.97          & 0.72 ± 0.06          & 0.52 ± 0.03          & 0.78 ± 0.02          & 0.79 ± 0.07          & 0.64 ± 0.04          & 1.11 ± 0.20           \\
CutMix                & 58.27 ± 0.06          & 56.31 ± 2.57          & 40.76 ± 1.70          & 55.94 ± 2.06          & 65.37 ± 0.56          & 64.69 ± 1.53          & 0.64 ± 0.02          & 0.56 ± 0.05          & 0.78 ± 0.05          & 0.75 ± 0.05          & 0.62 ± 0.03          & 0.58 ± 0.01           \\
PuzzleMix             & 59.80 ± 3.75          & 55.99 ± 1.48          & 39.16 ± 1.49          & 56.05 ± 3.34          & 66.00 ± 1.00          & 64.72 ± 2.97          & \textbf{0.59 ± 0.00} & 0.53 ± 0.02          & 0.81 ± 0.03          & 0.78 ± 0.04          & 0.62 ± 0.01          & 0.58 ± 0.08           \\
CoMix                 & 58.90 ± 2.14          & 58.58 ± 3.12          & 40.78 ± 1.68          & 54.04 ± 2.97          & 65.95 ± 1.74          & 64.39 ± 0.56          & 0.64 ± 0.00          & 0.50 ± 0.02          & 0.81 ± 0.03          & 0.82 ± 0.07          & 0.62 ± 0.03          & 0.58 ± 0.03           \\
GuidedMix             & 56.33 ± 0.06          & 70.22 ± 2.02          & 38.83 ± 1.69          & 53.07 ± 2.02          & 64.73 ± 1.48          & 64.07 ± 2.57          & 0.73 ± 0.01          & 0.33 ± 0.02          & 0.84 ± 0.04          & 0.85 ± 0.04          & 0.68 ± 0.03          & 0.61 ± 0.05           \\
DiffuseMix            & 58.27 ± 0.06          & 66.64 ± 2.03          & 40.77 ± 0.96          & 52.08 ± 1.14          & 65.70 ± 1.48          & 63.75 ± 1.12          & 0.69 ± 0.01          & 0.35 ± 0.03          & 0.83 ± 0.05          & 0.95 ± 0.05          & 0.62 ± 0.02          & 0.66 ± 0.02           \\
\pname{}              & \textbf{60.53 ± 1.99} & \textbf{73.46 ± 3.41} & \textbf{41.95 ± 0.35} & \textbf{56.31 ± 0.98} & \textbf{66.34 ± 1.12} & \textbf{66.02 ± 0.98} & 0.66 ± 0.01          & \textbf{0.29 ± 0.06} & \textbf{0.77 ± 0.02} & \textbf{0.73 ± 0.03} & \textbf{0.58 ± 0.02} & \textbf{0.57 ± 0.02}  \\ \hline
\multicolumn{1}{r}{~} & \multicolumn{12}{c}{~ ~ ~ ~ ~ ~ ~ KOA}                                                                                                                                                                                                                                                                                                \\ \hline
Mixup                 & 68.40 ± 0.26          & 76.89 ± 0.89          & 29.57 ± 25.23         & 63.96 ± 0.55          & 67.92 ± 0.26          & 67.16 ± 0.84          & 0.41 ± 0.01          & 0.23 ± 0.01          & 0.61 ± 0.02          & \textbf{0.49 ± 0.00} & 0.43 ± 0.01          & 0.42 ± 0.01           \\
CutMix                & 68.20 ± 0.26          & 76.75 ± 0.26          & 44.40 ± 0.55          & 62.90 ± 0.24          & 69.03 ± 0.57          & 67.59 ± 0.24          & \textbf{0.38 ± 0.00} & 0.23 ± 0.00          & 0.61 ± 0.02          & 0.51 ± 0.00          & 0.43 ± 0.02          & 0.42 ± 0.00           \\
PuzzleMix             & 68.37 ± 0.57          & 77.17 ± 0.10          & 43.56 ± 0.97          & 62.90 ± 0.23          & 69.18 ± 0.51          & 67.02 ± 0.21          & 0.40 ± 0.01          & 0.23 ± 0.00          & 0.63 ± 0.07          & 0.51 ± 0.01          & 0.41 ± 0.02          & 0.42 ± 0.01           \\
CoMix                 & 68.93 ± 0.75          & 77.21 ± 0.53          & 44.24 ± 0.49          & \textbf{63.58 ± 0.90} & 68.64 ± 0.24          & 67.79 ± 0.53          & 0.39 ± 0.03          & 0.23 ± 0.01          & 0.62 ± 0.04          & 0.49 ± 0.02          & 0.41 ± 0.01          & \textbf{0.40 ± 0.02}  \\
GuidedMix             & 68.07 ± 1.14          & 76.91 ± 0.15          & 44.40 ± 0.23          & 63.36 ± 0.35          & 68.92 ± 0.31          & 67.57 ± 0.43          & 0.41 ± 0.01          & 0.23 ± 0.00          & 0.61 ± 0.02          & 0.50 ± 0.01          & 0.43 ± 0.03          & 0.42 ± 0.01           \\
DiffuseMix            & 68.40 ± 0.44          & 77.40 ± 0.46          & 43.98 ± 0.18          & 63.54 ± 0.43          & 68.68 ± 0.31          & 67.64 ± 0.98          & 0.39 ± 0.01          & 0.23 ± 0.00          & 0.60 ± 0.03          & 0.49 ± 0.01          & 0.43 ± 0.02          & 0.41 ± 0.01           \\
\pname{}              & \textbf{69.80 ± 1.04} & \textbf{77.72 ± 0.16} & \textbf{44.53 ± 0.25} & 63.28 ± 0.76          & \textbf{69.41 ± 0.45} & \textbf{68.41 ± 0.36} & 0.40 ± 0.01          & \textbf{0.22 ± 0.00} & \textbf{0.57 ± 0.05} & 0.49 ± 0.02          & \textbf{0.40 ± 0.02} & 0.41 ± 0.01           \\ \hline
\multicolumn{1}{r}{~} & \multicolumn{12}{c}{~ ~ ~ ~ ~ ~ ~ CHAOYANG}                                                                                                                                                                                                                                                                                         \\ \hline
Mixup                 & 82.67 ± 0.65          & 88.29 ± 0.40          & 57.73 ± 1.20          & 82.53 ± 0.51          & 80.38 ± 2.66          & 85.36 ± 0.13          & 0.24 ± 0.01          & 0.13 ± 0.00          & 0.46 ± 0.01          & 0.25 ± 0.01          & 0.26 ± 0.02          & 0.21 ± 0.00           \\
CutMix                & 82.40 ± 0.30          & 88.21 ± 0.27          & 56.64 ± 0.51          & 82.12 ± 0.64          & 53.06 ± 3.56          & 84.91 ± 0.10          & 0.24 ± 0.00          & 0.13 ± 0.00          & 0.48 ± 0.01          & 0.26 ± 0.01          & 0.86 ± 0.04          & 0.22 ± 0.00           \\
PuzzleMix             & 83.17 ± 0.35          & 87.95 ± 0.35          & 57.71 ± 0.59          & 81.72 ± 0.57          & 54.81 ± 2.02          & 85.47 ± 0.49          & 0.23 ± 0.00          & 0.13 ± 0.00          & 0.47 ± 0.02          & 0.27 ± 0.01          & 0.84 ± 0.02          & 0.21 ± 0.01           \\
CoMix                 & 83.43 ± 0.25          & 88.10 ± 0.41          & 57.52 ± 0.71          & 81.84 ± 0.17          & 55.74 ± 3.51          & 85.18 ± 1.97          & 0.23 ± 0.01          & 0.13 ± 0.01          & 0.47 ± 0.00          & 0.26 ± 0.00          & 0.89 ± 0.02          & 0.21 ± 0.03           \\
GuidedMix             & 83.13 ± 0.35          & 87.72 ± 0.38          & 56.80 ± 0.33          & 81.64 ± 0.10          & 57.95 ± 1.90          & \textbf{85.72 ± 0.33} & 0.23 ± 0.01          & 0.13 ± 0.00          & 0.47 ± 0.01          & 0.27 ± 0.00          & 0.83 ± 0.01          & \textbf{0.21 ± 0.00}  \\
DiffuseMix            & 82.23 ± 0.06          & 87.56 ± 0.14          & 57.57 ± 0.45          & 80.87 ± 0.35          & 83.67 ± 0.51          & 85.39 ± 0.48          & 0.24 ± 0.01          & 0.13 ± 0.00          & \textbf{0.46 ± 0.00} & 0.28 ± 0.02          & \textbf{0.25 ± 0.01} & 0.21 ± 0.01           \\
\pname{}              & \textbf{84.10 ± 0.52} & \textbf{88.58 ± 0.26} & \textbf{58.06 ± 0.19} & \textbf{83.30 ± 0.45} & \textbf{84.27 ± 0.17} & 84.88 ± 0.66          & \textbf{0.22 ± 0.01} & \textbf{0.12 ± 0.00} & 0.47 ± 0.00          & \textbf{0.24 ± 0.01} & 0.26 ± 0.00          & 0.22 ± 0.01           \\ \hline
\multicolumn{1}{r}{~} & \multicolumn{12}{c}{~ ~ ~ ~ ~ ~ ~ GLEASON}                                                                                                                                                                                                                                                                                           \\ \hline
Mixup                 & 77.40 ± 0.62          & 93.96 ± 0.60          & 79.68 ± 6.27          & 87.78 ± 2.07          & 92.46 ± 0.57          & 88.93 ± 1.92         & 0.32 ± 0.01          & 0.05 ± 0.00          & 0.58 ± 0.29          & 0.17 ± 0.02          & 0.24 ± 0.06          & 0.15 ± 0.02           \\
CutMix                & 83.30 ± 4.15          & 94.88 ± 0.05          & \textbf{84.19 ± 2.96} & 88.99 ± 1.01          & 92.65 ± 0.31          & 89.09 ± 0.65         & 0.33 ± 0.05          & 0.05 ± 0.00          & \textbf{0.21 ± 0.05} & 0.16 ± 0.02          & 0.28 ± 0.02          & 0.16 ± 0.02           \\
PuzzleMix             & 81.43 ± 1.01          & 95.21 ± 0.39          & 71.60 ± 5.95          & 88.31 ± 0.45          & 92.23 ± 0.28          & 88.95 ± 1.73         & 0.33 ± 0.04          & 0.05 ± 0.00          & 0.22 ± 0.01          & 0.17 ± 0.01          & \textbf{0.24 ± 0.02} & 0.17 ± 0.03           \\
CoMix                 & 83.47 ± 1.67          & 94.77 ± 0.43          & 67.78 ± 1.43          & 88.88 ± 1.38          & 92.44 ± 0.28          & 88.45 ± 0.82         & 0.31 ± 0.05          & 0.06 ± 0.01          & 0.65 ± 0.10          & 0.16 ± 0.00          & 0.26 ± 0.02          & 0.17 ± 0.01           \\
GuidedMix             & 84.10 ± 2.44          & 95.05 ± 0.45          & 78.57 ± 5.04          & 88.87 ± 0.45          & 92.25 ± 0.29          & 88.64 ± 0.77         & 0.34 ± 0.01          & 0.05 ± 0.01          & 0.25 ± 0.04          & 0.17 ± 0.01          & 0.26 ± 0.05          & 0.17 ± 0.01           \\
DiffuseMix            & 85.63 ± 3.25          & \textbf{95.28 ± 0.36} & 77.94 ± 11.6         & 88.81 ± 0.81          & 92.74 ± 0.72          & 89.79 ± 1.78          & 0.33 ± 0.04          & 0.05 ± 0.00          & 0.46 ± 0.15          & 0.16 ± 0.02          & 0.28 ± 0.03          & 0.15 ± 0.02           \\
\pname{}              & \textbf{86.30 ± 3.84} & 94.77 ± 0.23          & 76.89 ± 8.60          & \textbf{89.22 ± 0.42} & \textbf{93.26 ± 1.15} & \textbf{90.77 ± 0.93} & \textbf{0.30 ± 0.03} & \textbf{0.04 ± 0.01} & 0.23 ± 0.01          & \textbf{0.15 ± 0.01} & 0.27 ± 0.06          & \textbf{0.14 ± 0.02}  \\ \hline
\end{tabular}
}
\captionof{table}{
Comparing \pname{} against six image mixup baselines—Mixup, CutMix, PuzzleMix, CoMix, GuidedMix, DiffuseMix
across six ordinal classification methods—POE, GOL, MWR, RnC, ORD2SEQ, CLOC
using four datasets—IDRID, KOA, Chaoyang and Gleason.
}
\label{table:main}
\end{table*}
\section{Experiments}
\subsubsection{Datasets And Preparation.}
\label{sec:data_setup}
The \textbf{Knee Osteoarthritis (KOA)} dataset~\cite{ds_koa} includes knee X-ray images (6,604 train/1,656 test) graded on the Kellgren–Lawrence (KL) scale (0–4).
The \textbf{Indian Diabetic Retinopathy (IDRID)}~\cite{ds_idrid} contains retinal fundus images (413/103) graded on the DR scale (0–4).
The \textbf{Chaoyang} dataset~\cite{ds_chaoyang} has colorectal histopathology images (4,021/2,139) graded by three pathologists, with $\sim$40\% disagreement in training and consensus test labels.
The \textbf{Crowd Gleason} dataset \cite{ds_gleason} contains prostate histopathology patches (16,151/2,926) graded by seven residents (mean $\kappa{=}0.54$) on the Gleason scale (NC, G3–G5), with consensus test labels. More in App.~\ref{app:datasets}.

Images are resized to $256{\times}256$, normalized to $[-1,1]$, and augmented with random flips and rotations $<20^{\circ}$; with histogram equalization for KOA.
IDRID and KOA simulate five labelers following~\cite{cloc}, randomly flipping 60\% grade 1$\to$2 and 20\% grade 2$\to$1, resulting in an overall 30\% grading variability in the dataset.
Chaoyang and Gleason contain real grading variability.
Each training image was assigned an ordinal soft label for \pname{} and a generic soft label for downstream classifiers (POE, GOL, etc.) during training, while test labels remained fixed.

\subsubsection{\pname{} Setup.}
We adopt the encoder, decoder, and quantizer architectures of~\cite{lat_diffusion} with pretrained \texttt{vq-f8} weights.
The non-ordinal and ordinal codebooks use $16{,}384$ and $512$, 4-dim entries, respectively, following pretrained compatibility and the smaller ordinal subspace~\cite{dis_aug}.
The discriminator implements~\cite{lat_diffusion}, with a linear classifier attached before the real/fake convolution for rank prediction.
\pname{} is trained for 400 epochs using Adam ($1{\times}10^{-5}$, batch size 11), freezing encoder downsampling blocks.
The checkpoint that maximizes mixing accuracy, measuring agreement between discriminator-predicted ranks and ranks expected from random latent interpolation, is selected for downstream tasks.

\subsubsection{Downstream Ordinal Model Setup.}
Ordinal baselines use their original implementation, extending loss functions to support generic soft labels where necessary.
Models are trained for 200 epochs using Adam (batch size 64), with 15\% of the training set reserved for validation and the best validation checkpoint used for testing.

\subsubsection{Baselines.}
We evaluate on six ordinal classification methods (POE\cite{poe}, GOL\cite{gol}, MWR\cite{mwr}, RnC\cite{rnc}, ORD2SEQ\cite{ord2seq}, and CLOC\cite{cloc}) across four datasets, comparing \pname{} with six state-of-the-art (SOTA) mixup methods: Mixup\cite{mixup}, CutMix\cite{cutmix}, PuzzleMix\cite{puzzlemix}, CoMix\cite{comix}, GuidedMix\cite{guidedmix}, and DiffuseMix\cite{diffusemix}.

\subsubsection{Evaluation Criteria.}
We report the mean and standard deviation of Accuracy and MAE over three runs, using a mixing probability of 0.5 unless otherwise specified.
\pname{} is evaluated by comparing the gains of ordinal algorithms when paired with \pname{} versus alternative mixup methods.

\subsubsection{Comparison with SOTA.}
Table~\ref{table:main} shows that,
\pname{} achieves the best mean accuracy and MAE in 20 and 15 of 24 settings, each defined by applying \pname{} with an ordinal algorithm on a dataset.
Compared with the strongest baseline, \pname{} significantly improves accuracy (one-sided Wilcoxon signed-rank test, $p=0.0075$) while typically reducing MAE by 4--6\%,  whereas other mixups often trade accuracy for MAE.
Surpassing diffusion-based methods like DiffuseMix underscores the benefit of explicitly modeling ordinal features.

\begin{figure}[t]
\centering
\includegraphics[width=\linewidth]{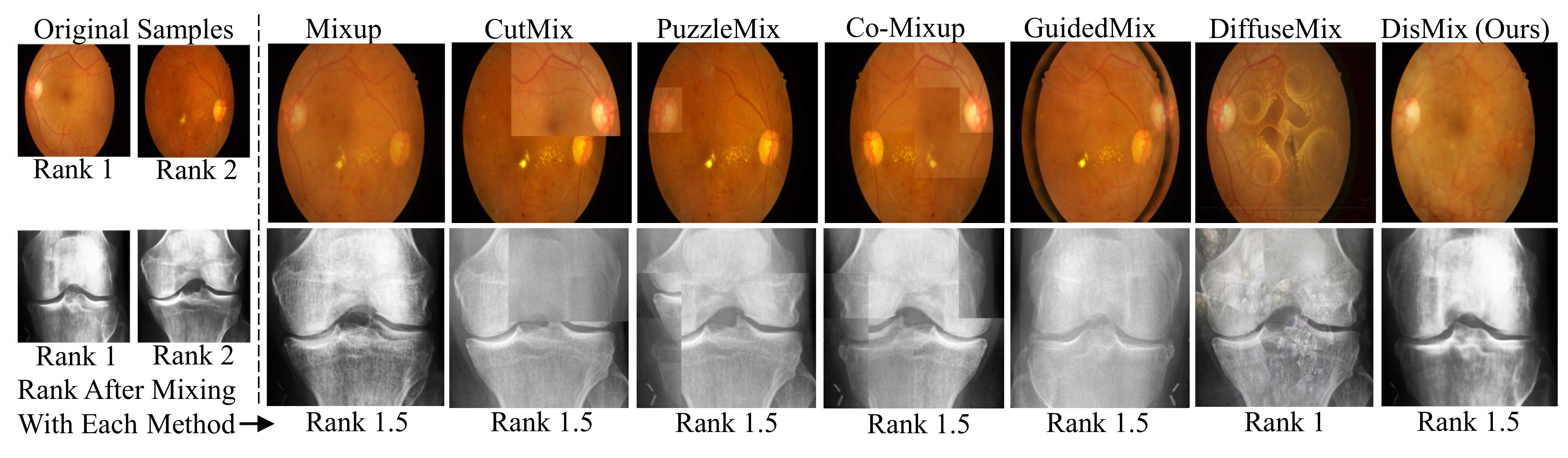}
\captionof{figure}{
Mixing a severe (rank 2) sample into a mild (rank 1) sample from IDRID (top) and KOA (bottom).
\pname{} produces plausible ordinal transitions from mild samples—such as darker lesion regions in IDRID and reduced bone gap in KOA—while preserving non-ordinal attributes like eye orientation and bone structure.
In contrast, other mixup often generate unrealistic structures, including multiple retinal optic discs, disoriented bone gaps, and imaging artifacts.
}
\label{fig:mixed_samples}
\end{figure}

Figure~\ref{fig:mixed_samples} compares mixed samples generated by different methods on IDRID and KOA.
Existing methods often produce unrealistic outputs, including duplicate optic discs and abnormal anatomical structures.
In contrast, \pname{} generates plausible severity transitions, such as darker lesions and narrower joint spaces, while preserving non-ordinal attributes including eye orientation and bone structure, demonstrating the benefit of mixing in disentangled latent spaces.

\subsubsection{Boundary Error Reduction.}
Beyond accuracy and MAE, we evaluate boundary-level classification errors using CLOC across datasets.
Figure~\ref{fig:boundary_errors} shows that \pname{} consistently achieves the lowest boundary errors, notably reducing 
IDRID C2$\leftrightarrow$C3 by 9\%,
Chaoyang C1$\leftrightarrow$C2 by 4\%,
Gleason C2$\leftrightarrow$C3 6\%
over to the best mixup baseline.

\subsubsection{Performance with Scarce Data.}
\label{sec:data_scarcity}
We restrict the training set to 10\% of samples per class (minimum 10 images) for both \pname{} and CLOC.
Table~\ref{table:scarcity} shows that \pname{} outperforms baseline mixup methods. 
Notably, on IDRID, \pname{} improves accuracy by ${\approx}1.5\%$ over the best baseline with only 60 original training images, demonstrating strong data efficiency and robustness in low-data regimes. More in App.~\ref{app:extended_results}.

\subsubsection{Robustness to Grading Variability.}
Following \cite{cloc}, we increase the boundary grade variability in IDRID and KOA by flipping 70/40\% and 90/60\% of grade 1$\to$2 / 2$\to$1 (yielding two settings with 50\% and 70\% overall grading variability for each dataset), and vary the mixup probability to 0.3, 0.5, and 0.7 with CLOC.
Tab.~\ref{table:noise_vs_mixprob} shows \pname{} performs best overall with a mix probability of 0.5, whereas 0.3 is preferable under mild grading variability (30\%) by providing sufficient data diversification.
Under severe variability (70\%), accuracy drops and MAE increases, especially at 0.7, indicating excessive mixing amplifies boundary-level variability.
More in App.~\ref{app:extended_results}.

\begin{table}[t]
\centering
\begin{minipage}[t]{0.46\linewidth}
\centering
\resizebox{\linewidth}{!}{%

\begin{tabular}{r|cc} \hline
\multicolumn{1}{l|}{~} & Accuracy $\uparrow$ & MAE $\downarrow$ \\ \hline \hline

\multicolumn{1}{l}{~} & \multicolumn{2}{c}{~ ~ ~ IDRID} \\ \hline
Mixup      & 43.36 $\pm$ 1.49 & 1.06 $\pm$ 0.04 \\
GuidedMix  & 46.09 $\pm$ 1.77 & 1.00 $\pm$ 0.09 \\
DiffuseMix & 45.26 $\pm$ 2.84 & 1.13 $\pm$ 0.08 \\
\pname{}   & \textbf{47.91 $\pm$ 2.82} & \textbf{0.99 $\pm$ 0.09} \\ \hline



\multicolumn{1}{l}{~} & \multicolumn{2}{c}{~ ~ ~ GLEASON} \\ \hline
Mixup      & 84.82 $\pm$ 0.95 & 0.21 $\pm$ 0.02 \\
GuidedMix  & 84.22 $\pm$ 0.22 & 0.23 $\pm$ 0.01 \\
DiffuseMix & 84.67 $\pm$ 1.46 & 0.23 $\pm$ 0.01 \\
\pname{}   & \textbf{85.38 $\pm$ 0.76} & \textbf{0.21 $\pm$ 0.01} \\ \hline
\end{tabular}
}
\captionof{table}{Performance under data scarcity, training set limited to 10\% of samples/class ($\ge$10 images/class).}
\label{table:scarcity}
\end{minipage}
\hfill
\begin{minipage}[t]{0.52\linewidth}
\centering
\resizebox{\linewidth}{!}{%

\begin{tabular}{c|ccc|ccc} \hline
\multirow{2}{*}{\begin{tabular}[c]{@{}c@{}}Vari.\\Level\end{tabular}}
& \multicolumn{3}{c|}{\multirow{2}{*}{Accuracy $\uparrow$}}
& \multicolumn{3}{c}{\multirow{2}{*}{MAE $\downarrow$}} \\
& \multicolumn{3}{c|}{}
& \multicolumn{3}{c}{} \\ \hline \hline

\multicolumn{1}{l}{~}
& \multicolumn{6}{c}{~ ~ ~ IDRID} \\ \hline

70\%
& {\cellcolor[rgb]{0.922,0.969,0.933}}56.99
& {\cellcolor[rgb]{0.831,0.929,0.855}}58.89
& 55.33
& {\cellcolor[rgb]{0.961,0.749,0.635}}0.67
& {\cellcolor[rgb]{0.965,0.784,0.686}}0.65
& {\cellcolor[rgb]{0.945,0.663,0.514}}0.72 \\

50\%
& {\cellcolor[rgb]{0.518,0.8,0.592}}65.37
& {\cellcolor[rgb]{0.643,0.851,0.698}}62.78
& {\cellcolor[rgb]{0.388,0.745,0.482}}67.98
& {\cellcolor[rgb]{0.988,0.918,0.882}}0.57
& {\cellcolor[rgb]{0.976,0.851,0.784}}0.61
& 0.52 \\

30\%
& {\cellcolor[rgb]{0.443,0.769,0.529}}66.92
& {\cellcolor[rgb]{0.486,0.788,0.565}}66.02
& {\cellcolor[rgb]{0.549,0.812,0.616}}64.73
& {\cellcolor[rgb]{0.988,0.918,0.882}}0.57
& {\cellcolor[rgb]{0.988,0.918,0.882}}0.57
& {\cellcolor[rgb]{0.969,0.8,0.71}}0.64 \\ \hline

\multicolumn{1}{l}{~}
& \multicolumn{6}{c}{~ ~ ~ KOA} \\ \hline

70\%
& 64.46
& {\cellcolor[rgb]{0.835,0.933,0.863}}65.57
& {\cellcolor[rgb]{0.945,0.976,0.953}}64.85
& {\cellcolor[rgb]{0.945,0.663,0.514}}0.47
& {\cellcolor[rgb]{0.965,0.776,0.678}}0.45
& {\cellcolor[rgb]{0.957,0.722,0.596}}0.46 \\

50\%
& {\cellcolor[rgb]{0.482,0.784,0.565}}67.93
& {\cellcolor[rgb]{0.443,0.769,0.529}}68.19
& {\cellcolor[rgb]{0.388,0.745,0.482}}68.55
& {\cellcolor[rgb]{0.992,0.945,0.922}}0.42
& 0.41
& 0.41 \\

30\%
& {\cellcolor[rgb]{0.467,0.78,0.549}}68.03
& {\cellcolor[rgb]{0.412,0.757,0.502}}68.41
& {\cellcolor[rgb]{0.459,0.776,0.541}}68.09
& {\cellcolor[rgb]{0.992,0.945,0.922}}0.42
& 0.41
& 0.41 \\ \hline

\begin{tabular}[c]{@{}c@{}}Mix\\Prob.\end{tabular}
& 0.3 & 0.5 & 0.7
& 0.3 & 0.5 & 0.7 \\ \hline

\end{tabular}
}
\captionof{table}{\pname{}'s performance heatmaps under grading variability (30\%, 50\%, 70\%) and mix probabilities (0.3, 0.5, 0.7).}
\label{table:noise_vs_mixprob}
\end{minipage}
\end{table}
\begin{figure}[tb]
    \centering
    \includegraphics[width=0.95\linewidth]{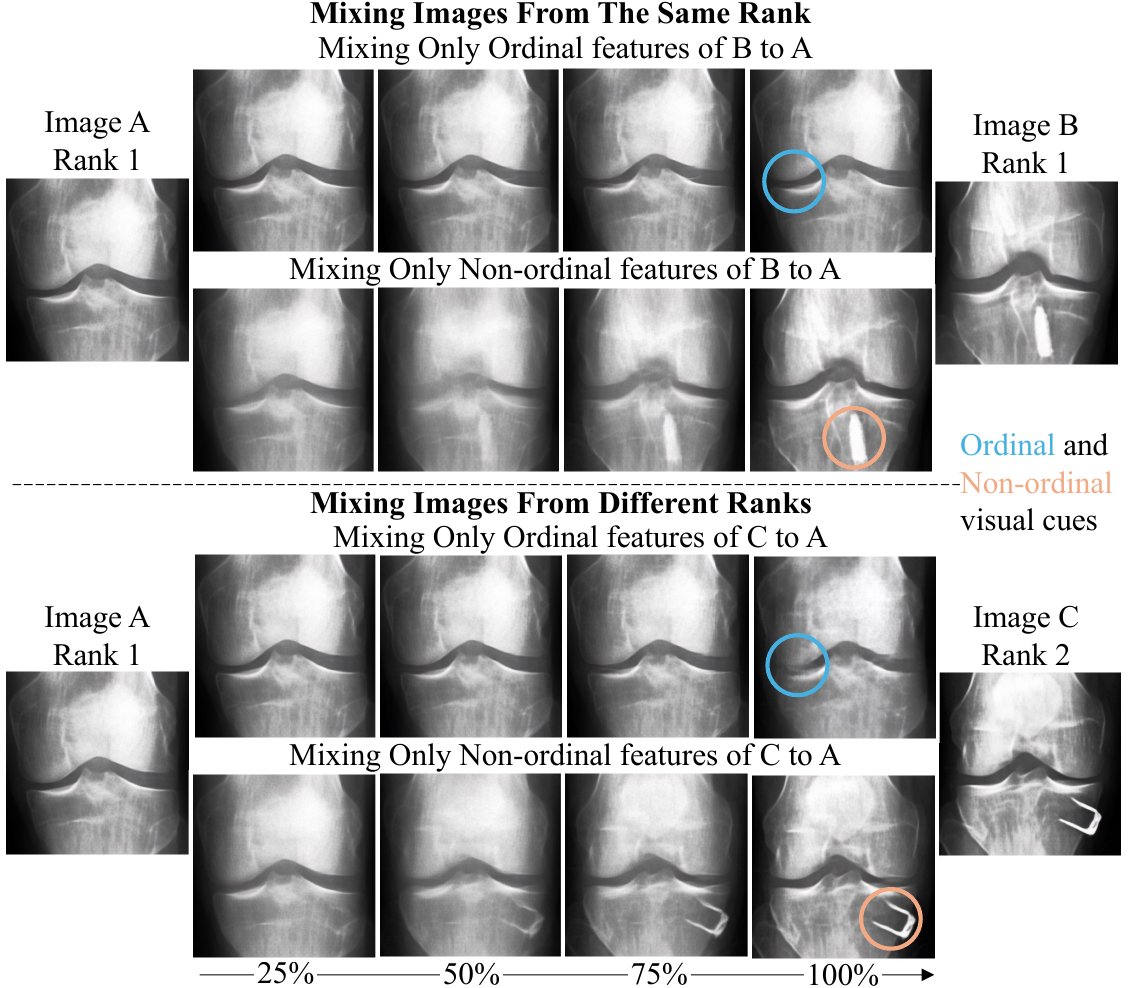}
    \captionof{figure}{Mixing either Ordinal or Non-ordinal features into the reference image (A) from the same rank (B) and severe rank (C) samples from KOA dataset according to the percentage in x-axis.
    Mixing ordinal features from (B) preserves the joint space, whereas mixing from (C) narrows it.
    In contrast, mixing non-ordinal features primarily changes visual appearance, including attributes such as nails.}
    \label{fig:order_transition}
\end{figure}

\section{Ablation Studies and Discussion}

\begin{figure*}[tb]
    \centering
    \begin{minipage}[]{0.28\textwidth}
        \centering
        \resizebox{\linewidth}{!}{%
        
\begin{tabular}{l|cc|cc} 
\hline
~                       & \multicolumn{2}{c|}{Accuracy $\uparrow$} & \multicolumn{2}{c}{MAE $\downarrow$}   \\ 
\hline
~                       & on $z^o$ & on $z^n$           & on $z^o$ & on $z^n$     \\ 
\hline\hline
DisMix                  & 65.04     & 19.65              & 0.45      & 1.42          \\
Without GRL             & 62.13     & 22.95              & 0.48      & 1.36          \\
Without $\ell_{\text{orth}}$            & 59.22     & 24.35              & 0.59      & 1.38          \\
One codebook            & \multicolumn{2}{c|}{33.01}    & \multicolumn{2}{c}{1.72}  \\
\hline
\end{tabular}

        }
        \captionof{table}{Probe classifier performance with \pname{} variants.}
        \label{table:quantitative_disent}
        \resizebox{\linewidth}{!}{%
        
\begin{tabular}{r|cc} \hline
\multicolumn{1}{l|}{~} & Accuracy $\uparrow$ & MAE $\downarrow$ \\ \hline \hline

\multicolumn{1}{l}{~} & \multicolumn{2}{c}{~ ~ ~ IDRID} \\ \hline
Ordinal Mix     & 65.09 $\pm$ 1.73 & 0.56 $\pm$ 0.02 \\
Generate \& Mix & 67.01 $\pm$ 2.56 & 0.55 $\pm$ 0.04 \\
Non-Ordinal Mix & 64.74 $\pm$ 1.17 & 0.60 $\pm$ 0.01 \\
Order Swap      & 66.69 $\pm$ 2.96 & 0.54 $\pm$ 0.06 \\ \hline

\multicolumn{1}{l}{~} & \multicolumn{2}{c}{~ ~ ~ GLEASON} \\ \hline
Ordinal Mix     & 88.96 $\pm$ 1.31 & 0.17 $\pm$ 0.01 \\
Generate \& Mix & 90.34 $\pm$ 1.43 & 0.13 $\pm$ 0.02 \\
Non-Ordinal Mix & 89.62 $\pm$ 1.19 & 0.16 $\pm$ 0.02 \\
Order Swap      & 90.39 $\pm$ 0.97 & 0.14 $\pm$ 0.02 \\ \hline
\end{tabular}
        }
        \captionof{table}{Performance vs. individual mixing policies with CLOC.}
        \label{table:abl_mix_policies}
    \end{minipage}
    \hfill
    \begin{minipage}[]{0.42\textwidth}
        \centering
        \includegraphics[width=0.94\linewidth]{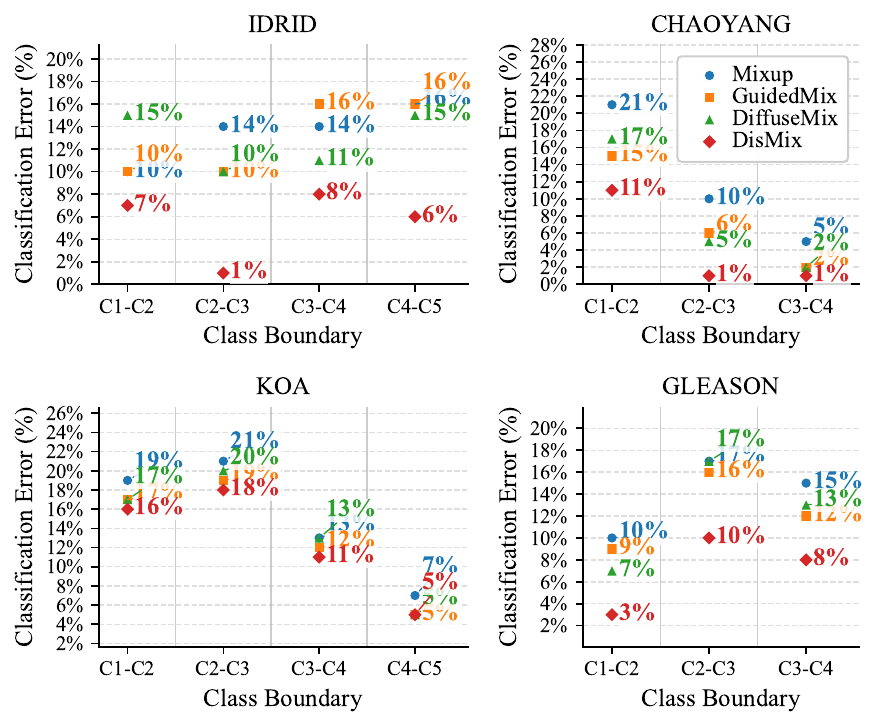}
        \caption{Per-boundary classification error rates of \pname{} vs. baseline mixup methods.}
        \label{fig:boundary_errors}
    \end{minipage}
    \hfill
    \begin{minipage}[]{0.27\textwidth}
        \centering
        \resizebox{\linewidth}{!}{%
        
\begin{tabular}{r|cc} \hline
\multicolumn{1}{c|}{~} & Accuracy $\uparrow$ & MAE $\downarrow$ \\ \hline \hline

\multicolumn{1}{c}{~} & \multicolumn{2}{c}{~ ~ IDRID} \\ \hline
Mixup      & 65.37 $\pm$ 2.25 & 0.58 $\pm$ 0.02 \\
CutMix     & 64.39 $\pm$ 0.56 & 0.59 $\pm$ 0.06 \\
PuzzleMix  & 65.69 $\pm$ 0.56 & 0.56 $\pm$ 0.00 \\
CoMix      & 65.04 $\pm$ 0.00 & 0.56 $\pm$ 0.02 \\
GuidedMix  & 64.72 $\pm$ 1.48 & 0.56 $\pm$ 0.01 \\
DiffuseMix & 65.69 $\pm$ 2.02 & 0.58 $\pm$ 0.06 \\
\pname{}   & \textbf{66.01 $\pm$ 0.98} & \textbf{0.55 $\pm$ 0.03} \\ \hline

\multicolumn{1}{c}{~} & \multicolumn{2}{c}{~ ~ KOA} \\ \hline
Mixup      & 68.55 $\pm$ 0.37 & 0.39 $\pm$ 0.01 \\
CutMix     & 68.97 $\pm$ 0.82 & 0.38 $\pm$ 0.02 \\
PuzzleMix  & 68.45 $\pm$ 0.35 & 0.38 $\pm$ 0.02 \\
CoMix      & 68.37 $\pm$ 0.69 & 0.39 $\pm$ 0.01 \\
GuidedMix  & 68.49 $\pm$ 0.09 & 0.39 $\pm$ 0.01 \\
DiffuseMix & 68.35 $\pm$ 0.38 & 0.39 $\pm$ 0.02 \\
\pname{}   & \textbf{69.53 $\pm$ 0.47} & \textbf{0.38 $\pm$ 0.01} \\ \hline
\end{tabular}

        }
        \captionof{table}{\pname{} vs. mixup baselines with CLOC with no grading variability.}
        \label{table:clean_main}
    \end{minipage}
\end{figure*}

\subsubsection{Quality of Disentanglement.}
The rightmost panel of Fig.\ref{fig:teaser} shows the disentanglement on KOA, with the ordinal codebook exhibiting a rank-aligned structure and the non-ordinal codebook no apparent rank organization.
We quantify this by training linear probes on the frozen latent features ($z^o$ and $z^n$) and reporting test-set rank prediction accuracy.
As shown in Tab.\ref{table:quantitative_disent}, the probe achieves 65.04\% on $z^o$ vs. 19.65\% on $z^n$, confirming that ordinal information is concentrated in $z^o$, while $z^n$ remains at chance for five-class KOA.
An ablation study further shows that removing the GRL or $\ell_{\text{sim}}$ increases rank leakage into $z^n$, whereas removing the dual codebooks collapses accuracy to 33\%, validating the proposed design.

Figure~\ref{fig:order_transition} illustrates that the disentanglement captures clinically meaningful factors of knee osteoarthritis grading.
Joint-space narrowing, a key severity indicator, is encoded in the ordinal latent, whereas the surgical nail, unrelated to grading, is captured in the non-ordinal latent~\cite{koa_discussion1, koa_discussion2}.
Accordingly, modifying the non-ordinal latent alters the nail (rows 2 and 4), while modifying the ordinal latent changes the joint space (row 3), demonstrating a controllable and semantically meaningful decomposition.

\subsubsection{Analysis of DisMix Components and Policies.}
To assess the contribution of each component, we evaluate the proposed mixup policies individually. Ordinal Mix, Non-Ordinal Mix, Generate \& Mix, and Order Swap isolate the ordinal branch, non-ordinal branch, conditional rank generation, and learned feature exchange, respectively.
Table \ref{table:abl_mix_policies} shows, each policy is competitive against each other, but remains weaker than their combined use in Tab.\ref{table:main}, highlighting their complementary roles.
Tab.\ref{table:extreme_setting} (App.~\ref{app:extended_results}) finds that Generate \& Mix achieves higher accuracy in severe data scarcity (10\% images/class) and extreme grading variability (70\%) likely due to its two-stage process of generating a clean adjacent rank before mixing, highlighting the benefit of ordinal-conditioned synthesis. 
Tab.\ref{table:inter_vs_pres} (App.~\ref{app:extended_results}) further group policies based on label-preserving and label-interpolating and show that label-preserving policies slightly outperform interpolating policies when used individually.

\subsubsection{Without Grading Variability.}
Although \pname{} was evaluated primarily with grading variability, we also evaluate without grading variability (i.e. a single label per image).
Table~\ref{table:clean_main} shows that \pname{} achieves the best accuracy and MAE on both IDRID and KOA, indicating that \pname{} remains effective even in the absence of grading variability.

\subsubsection{Scope of Generations.}
Although Fig.~\ref{fig:mixed_samples} and \ref{fig:order_transition} show that \pname{} produces visually plausible ordinal and non-ordinal variations, the generated images are intended for data augmentation rather than direct clinical interpretation, as the generator prioritizes preserving ordinal semantics over high-fidelity image synthesis.
Furthermore, the interpolated ranks are used solely for soft supervision and should not be interpreted as new clinically defined grades.

\begin{figure}[tb]
    \centering
    \includegraphics[width=0.8\linewidth]{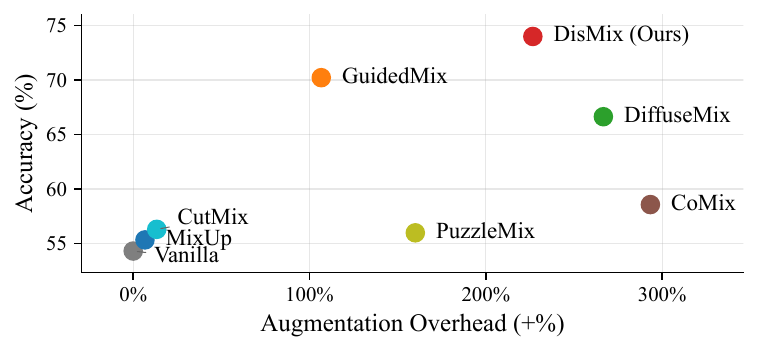}
    \caption{Accuracy vs. augmentation overhead of \pname{} vs. mixup baselines on IDRID.}
    \label{fig:aug_overhead}
\end{figure}

\subsubsection{Limitations.}
We measure mixup augmentation overhead as the percentage increase in downstream training time relative to vanilla training (without mixup) on IDRID using GOL.
Figure~\ref{fig:aug_overhead} shows that \pname{} offers a good accuracy--efficiency trade-off, outperforming mixup baselines while incurring lower overhead than DiffuseMix and Co-Mixup.
This can be further mitigated by generating and caching mixed samples offline before downstream training.
Furthermore, training \pname{} incurs a one-time, per-dataset overhead to learn the disentangled representation, which is amortized across all downstream classifiers (see App.~\ref{app:extended_results}).
Future work will explore foundation models for few-shot ordinal disentanglement to reduce per-dataset training.

\section{Conclusion}
We presented \pname{}, a latent mixup method for ordinal data motivated by the observation that conventional image mixup entangles order-relevant and order-irrelevant factors, thereby weakening the ordinal signal.
\pname{} disentangles ordinal and non-ordinal features into separate latent subspaces via a dual-codebook VQ-VAE, enabling order-aware mixing within dedicated subspaces.
Extensive experiments show that \pname{} achieves strongest aggregate performance among six SOTA mixup baselines across six ordinal classifiers, improving both accuracy and MAE.
Our results underscore the importance of order-aware mixup for ordinal data highlighting disentanglement-based methods as a promising direction for robust ordinal learning in medical imaging.

{\small
\bibliography{aaai2027}
}

\clearpage
\appendix
\section*{Appendix}

\section{Ordinal Soft Label}
\label{app:soft_ord_label}
This section elaborates on the ordinal soft-label creation process with a fail-safe mechanism and examples.

Each sample $x_i$, graded by more than one annotators with scalar ranks $\{y_{i,a}\}_{a \in \mathcal{A}}$,
is assigned a single ordinal soft label.
Firstly, each label ${y}_{i,a}$ is converted into a progressive vector 
$\mathbf{p}_{i,a}^{(k)} = \mathbb{I}(k < \tilde{y}_{i,a})$ for $k \in [0, C{-}2]$.
Secondly, taking the element-wise average across annotators yields
$\mathbf{s}_i^{(k)} {=} \tfrac{1}{|\mathcal{A}|}\sum_{a \in \mathcal{A}} p_{i,a}^{(k)}$, forming $\mathbf{s}_i {=} [s_i^{(0)}, \dots, s_i^{(C-2)}]$.
Lastly, to ensure a non-increasing progression, an optional 
monotonicity constraint is applied to $\mathbf{s}_i$,
ensuring $s_i^{(k+1)} \le s_i^{(k)}$ for all $k$. 
The resulting target $\ddot{y}_i$ reflects that belonging to a higher rank (e.g., 4) implies inclusion in all preceding ranks.
For instance, if $C{=}5$,  $|\mathcal{A}|{=}4$, and annotations,
\begin{itemize}
    \item $\{y_{i,a}\}_{a\in\mathcal{A}} {=} \{3, 3, 4, 4\} \to \ddot{y}_i {=} [1, 1, 1, 0.5]$
    \item $\{y_{i,a}\}_{a\in\mathcal{A}} {=} \{0, 1, 2, 2\} \to \ddot{y}_i {=} [0.75, 0.5, 0, 0]$
\end{itemize}
If $C{=}5$,  $|\mathcal{A}|{=}3$, and annotations,
\begin{itemize}
    \item $\{y_{i,a}\}_{a\in\mathcal{A}} {=} \{0, 1, 1\} \to \ddot{y}_i {=} [0.66,0,0,0]$
\end{itemize}

\section{Datasets}
\label{app:datasets}
\subsubsection{Knee Osteoarthritis (KOA) dataset.}
The training set includes 2,286, 1,046, 1,516, 757, and 173 x-ray images for grades 0--4.
The validation set contains 328, 153, 212, 106, and 27 respectively.
We combine the above into a single training set for \pname{} training.
The testing split has 639, 296, 447, 223, and 51 respectively.

\subsubsection{Indian Diabetic Retinopathy Image Dataset (IDRID).}
The training set contains 134, 20, 136, 74, and 49 fundus images for grades 0--4, respectively. The official test set includes 34, 5, 32, 19, and 13 images for grades 0--4.

\subsubsection{Chaoyang dataset.}
The training set contains 1,111 normal, 842 serrated, 1,404 adenocarcinoma, and 664 adenoma images.
The consensus-labeled test set includes 705, 321, 840, and 273 images, respectively.

\subsubsection{Crowd Gleason dataset.}
The majority-vote training set contains 7,974 non-cancer, 2,038 grade 3, 2,840 grade 4, and 972 grade 5 images.
The expert–resident consensus test set includes 2,157, 548, 164, and 57 images, respectively.

\section{Extended Results}
\label{app:extended_results}

\subsubsection{Extended Analysis with Data Scarcity.}
We extend the analysis under data scarcity to more datasets and methods in Tab.~\ref{table:scarcity_extended} and show that \pname{} performs well under data scarce conditions.
\begin{table}[tb]
\centering
\resizebox{\linewidth}{!}{%
\begin{tabular}{r|cc|cc} \hline
\multicolumn{1}{l|}{~} & \multicolumn{2}{c|}{Accuracy $\uparrow$}                 & \multicolumn{2}{c}{Mean Absolute Error $\downarrow$}      \\ \hline
\multicolumn{1}{c|}{~} & ORD2SEQ               & CLOC                  & ORD2SEQ              & CLOC                  \\ \hline\hline
\multicolumn{1}{l}{~}  & \multicolumn{4}{c}{ ~ ~ ~ IDRID}                                                                    \\ \hline
Mixup                  & 49.92 ± 1.06          & 43.36 ± 1.49          & 0.88 ± 0.05          & 1.06 ± 0.04           \\
GuidedMix              & 49.82 ± 2.41          & 46.09 ± 1.77          & 0.88 ± 0.08          & 1.00 ± 0.09           \\
DiffuseMix             & 48.22 ± 1.12          & 45.26 ± 2.84          & 0.93 ± 0.02          & 1.13 ± 0.08           \\
\pname{}                   & \textbf{51.44 ± 1.67} & \textbf{47.91 ± 2.82} & \textbf{0.86 ± 0.06} & \textbf{0.99 ± 0.09}  \\ \hline
\multicolumn{1}{l}{}   & \multicolumn{4}{c}{ ~ ~ KOA}                                                                      \\ \hline
Mixup                  & 58.22 ± 1.19          & 58.77 ± 1.03          & 0.58 ± 0.02          & 0.57 ± 0.02           \\
GuidedMix              & 58.01 ± 1.40          & 57.93 ± 0.07          & 0.60 ± 0.01          & 0.59 ± 0.03           \\
DiffuseMix             & 58.13 ± 1.14          & 59.00 ± 0.98          & 0.59 ± 0.02          & \textbf{0.56 ± 0.00}  \\
\pname{}                   & \textbf{58.58 ± 0.32} & \textbf{59.53 ± 1.40} & \textbf{0.58 ± 0.01} & 0.58 ± 0.04           \\ \hline
\multicolumn{1}{l}{}   & \multicolumn{4}{c}{ ~ ~ ~ CHAOYANG}                                                                 \\ \hline
Mixup                  & 78.37 ± 0.20          & 79.52 ± 0.42          & 0.33 ± 0.01          & 0.31 ± 0.00           \\
GuidedMix              & 78.99 ± 0.56          & \textbf{80.02 ± 0.60} & 0.33 ± 0.01          & \textbf{0.30 ± 0.01}  \\
DiffuseMix             & 78.20 ± 0.18          & 78.68 ± 0.74          & 0.33 ± 0.00          & 0.32 ± 0.02           \\
\pname{}                   & \textbf{79.38 ± 0.48} & 78.77 ± 0.70          & \textbf{0.32 ± 0.03} & 0.32 ± 0.01           \\ \hline
\multicolumn{1}{l}{}   & \multicolumn{4}{c}{ ~ ~ ~ GLEASON}                                                            \\ \hline
Mixup                  & 86.15 ± 1.19          & 84.82 ± 0.95          & 0.23 ± 0.01          & 0.21 ± 0.02           \\
GuidedMix              & 86.35 ± 0.71          & 84.22 ± 0.22          & 0.21 ± 0.00          & 0.23 ± 0.01           \\
DiffuseMix             & 87.07 ± 0.52          & 84.67 ± 1.46          & 0.23 ± 0.03          & 0.23 ± 0.01           \\
\pname{}                   & \textbf{87.54 ± 0.62} & \textbf{85.38 ± 0.76} & \textbf{0.20 ± 0.01} & \textbf{0.21 ± 0.01}  \\ \hline
\end{tabular}
}
\captionof{table}{Extended results of performance under data scarcity, with the training set limited to 10\% of samples per class ($\ge$10 images/class).}
\label{table:scarcity_extended}
\end{table}

\subsubsection{Extended Analysis of DisMix Components and Policies.}
We extend the analysis of DisMix components and policies to more datasets and ordinal methods in Tab.~\ref{table:abl_mix_policies_extended} and show that policies are competitive against each other but remains suboptimal to their combined use in Tab.~\ref{table:main}.
\begin{table}[tb]
\centering
\arrayrulecolor{black}
\resizebox{\linewidth}{!}{%
\begin{tabular}{r|cc|cc} \hline
\multicolumn{1}{l|}{~} & \multicolumn{2}{c|}{Accuracy $\uparrow$}              & \multicolumn{2}{c}{Mean Absolute Error $\downarrow$}                  \\ \hline
\multicolumn{1}{l|}{~} & ORD2SEQ      & CLOC         & ORD2SEQ     & CLOC         \\ \hline\hline
\multicolumn{1}{l}{~}  & \multicolumn{4}{c}{~ ~ ~ IDRID}                                                       \\ \hline
Ordinal Mix            & 66.67 ± 0.56 & 65.09 ± 1.73 & 0.61 ± 0.02 & 0.56 ± 0.02  \\
Generate \& Mix        & 64.73 ± 0.56 & 67.01 ± 2.56 & 0.62 ± 0.02 & 0.55 ± 0.04  \\
Non-Ordinal Mix        & 64.73 ± 2.02 & 64.74 ± 1.17 & 0.59 ± 0.05 & 0.60 ± 0.01  \\
Order Swap             & 65.37 ± 2.44 & 66.69 ± 2.96 & 0.65 ± 0.04 & 0.54 ± 0.06  \\ \hline
\multicolumn{1}{r}{~}  & \multicolumn{4}{c}{~ ~ ~ KOA}                                                         \\ \hline
Ordinal Mix            & 68.90 ± 0.27 & 68.35 ± 0.79 & 0.43 ± 0.01 & 0.41 ± 0.01  \\
Generate \& Mix        & 68.72 ± 0.22 & 68.27 ± 0.40 & 0.42 ± 0.02 & 0.41 ± 0.01  \\
Non-Ordinal Mix        & 68.36 ± 0.27 & 67.71 ± 0.33 & 0.42 ± 0.01 & 0.42 ± 0.01  \\
Order Swap             & 68.88 ± 0.48 & 67.93 ± 0.36 & 0.43 ± 0.00 & 0.41 ± 0.00  \\ \hline
\multicolumn{1}{r}{~}  & \multicolumn{4}{c}{~ ~ ~ CHAOYANG}                                                    \\ \hline
Ordinal Mix            & 84.29 ± 0.29 & 84.77 ± 0.39 & 0.27 ± 0.00 & 0.22 ± 0.01  \\
Generate \& Mix        & 84.14 ± 0.59 & 84.07 ± 0.71 & 0.26 ± 0.02 & 0.23 ± 0.02  \\
Non-Ordinal Mix        & 84.10 ± 0.13 & 84.27 ± 0.88 & 0.25 ± 0.01 & 0.23 ± 0.01  \\
Order Swap             & 83.54 ± 0.00 & 84.44 ± 1.11 & 0.26 ± 0.00 & 0.24 ± 0.03  \\ \hline
\multicolumn{1}{r}{~}  & \multicolumn{4}{c}{~ ~ ~ GLEASON}                                                     \\ \hline
Ordinal Mix            & 92.12 ± 0.67 & 88.96 ± 1.31 & 0.23 ± 0.04 & 0.17 ± 0.01  \\
Generate \& Mix        & 91.89 ± 0.24 & 90.34 ± 1.43 & 0.27 ± 0.07 & 0.13 ± 0.02  \\
Non-Ordinal Mix        & 91.81 ± 0.44 & 89.62 ± 1.19 & 0.26 ± 0.04 & 0.16 ± 0.02  \\
Order Swap             & 92.32 ± 0.55 & 90.39 ± 0.97 & 0.29 ± 0.02 & 0.14 ± 0.02  \\ \hline
\end{tabular}
}
\captionof{table}{Extended performance comparison of individual mixing policies.}
\label{table:abl_mix_policies_extended}
\end{table}

\subsubsection{Extended Analysis of Policy Behavior Under Extreme Conditions.}
We examine policy behavior under severe data scarcity (10\% per class) and extreme grading variability (70\% of boundary grade variability) using CLOC on KOA dataset.
As shown in Tab.\ref{table:extreme_setting}, Generate \& Mix achieves ${\approx}1{-}2\%$ higher accuracy than the respective combined-policies setting for CLOC in Tab.~\ref{table:scarcity_extended} (with data scarcity) and Tab.~\ref{table:noise_vs_mixprob} (70\% grading variability with 0.5 mix probability),
likely due to its two-stage process of generating a clean adjacent rank before mixing.
This highlights the advantage of ordinal-conditioned synthesis over interpolation between available samples under extreme conditions.
Order Swap, the second-best, indicates that label-preserving mixup remains effective even under extreme settings.
\begin{table}[tb]
\centering
\arrayrulecolor{black}
\resizebox{\linewidth}{!}{%
\begin{tabular}{r|cc|cc} \hline
\multicolumn{1}{l|}{~} & \multicolumn{2}{c!{\color{black}\vrule}}{Severe Scarcity} & \multicolumn{2}{c}{Extreme Grading Variability}             \\ \hline
\multicolumn{1}{l|}{~} & Accuracy $\uparrow$   & MAE $\downarrow$                   & Accuracy $\uparrow$   & MAE $\downarrow$      \\ \hline\hline
Ordinal Mix            & 59.75 ± 0.47          & 0.57 ± 0.00                        & 64.45 ± 2.11          & 0.45 ± 0.02           \\
Generate \& Mix        & \textbf{60.28 ± 1.13} & \textbf{0.56 ± 0.02}               & \textbf{66.57 ± 2.11} & \textbf{0.44 ± 0.02}  \\
Non-Ordinal Mix        & 59.60 ± 1.05          & 0.58 ± 0.02                        & 65.27 ± 1.47          & 0.46 ± 0.02           \\
Order Swap             & \textit{59.90 ± 0.27} & \textit{0.57 ± 0.02}               & \textit{65.88 ± 0.07} & \textit{0.45 ± 0.00}  \\ \hline
\end{tabular}
}
\captionof{table}{Performance of mixup policies in severe data scarcity (10\% data per class) and extreme grading variability (${\approx}70\%$) with CLOC on KOA dataset.}
\label{table:extreme_setting}
\arrayrulecolor{black}
\end{table}

\subsubsection{Extended Analysis of Label-Preserving vs. Interpolating Mixup.}
Table \ref{table:inter_vs_pres} groups policies based on label-preserving (Non-ordinal Mix and Order Swap) and label-interpolating (Ordinal Mix and Generate \& Mix) and show that label-preserving policies slightly outperform interpolating policies when used individually on KOA, Chaoyang and Gleason.
\begin{table}
\centering
\arrayrulecolor{black}
\resizebox{\linewidth}{!}{%
\begin{tabular}{r!{\color{black}\vrule}cc|cc} \hline
~                     & \multicolumn{2}{c|}{Accuracy $\uparrow$}              & \multicolumn{2}{c}{Mean Absolute Error $\downarrow$}                  \\ \hline
~                     & ORD2SEQ      & CLOC        & ORD2SEQ     & CLOC         \\ \hline\hline
\multicolumn{1}{r}{~} & \multicolumn{4}{c}{~ ~ IDRID}                                                             \\ \hline
Interpolating          & \textbf{65.70 ± 2.02} & \textbf{65.36 ± 0.56} & 0.65 ± 0.03 & 0.59 ± 0.02  \\
Preserving             & 64.08 ± 2.57 & 65.04 ± 0.00 & \textbf{0.62 ± 0.05} & \textbf{0.58 ± 0.02}  \\ \hline
\multicolumn{1}{r}{~} & \multicolumn{4}{c}{~ ~ KOA}                                                               \\ \hline
Interpolating         & 68.56 ± 0.18 & 67.89 ± 0.25 & 0.41 ± 0.01 & 0.42 ± 0.00  \\
Preserving            & \textbf{68.62 ± 0.17} & \textbf{68.13 ± 0.41} & \textbf{0.41 ± 0.00} & \textbf{0.41 ± 0.00}  \\ \hline
\multicolumn{1}{r}{~} & \multicolumn{4}{c}{~ ~ CHAOYANG}                                                          \\ \hline
Interpolating         & 83.79 ± 0.58 & 83.99 ± 0.59 & 0.28 ± 0.03 & 0.23 ± 0.01  \\
Preserving            & \textbf{83.82 ± 0.05} & \textbf{84.63 ± 0.41} & \textbf{0.27 ± 0.01} & \textbf{0.23 ± 0.00}  \\ \hline
\multicolumn{1}{r}{~} & \multicolumn{4}{c}{~ ~ GLEASON}                                                           \\ \hline
Interpolating         & 91.37 ± 0.59 & 88.85 ± 0.89 & \textbf{0.24 ± 0.09} & 0.17 ± 0.02  \\
Preserving            & \textbf{91.65 ± 0.54} & \textbf{89.86 ± 0.34} & 0.27 ± 0.05 & \textbf{0.16 ± 0.00}  \\ \hline
\end{tabular}
}
\captionof{table}{\pname{} performance with label-preserving vs. interpolating Mixups.}
\label{table:inter_vs_pres}
\end{table}

\subsubsection{\pname{} Training Overhead.}
We compare the training overhead of \pname{} and DiffuseMix on IDRID using a single RTX 4090 GPU. \pname{} trains in only $\approx$1.3 hours, over $6\times$ faster than DiffuseMix fine-tuning ($\approx$8.5 hours), while achieving higher accuracy and lower MAE as shown in Tab.~\ref{table:dismix_vs_diffusemix_overhead}.

\begin{table}[b]
\centering
\resizebox{0.5\linewidth}{!}{%
\begin{tabular}{r|c|c} 
\hline
\multicolumn{1}{c|}{~} & Accuracy $\uparrow$ & MAE $\downarrow$ \\ 
\hline\hline
Base   & 63.75 & 0.66 \\
FT     & 64.30 & 0.61 \\ \hline
DisMix & 66.02 & 0.57 \\
\hline
\end{tabular}
}
\caption{\pname{} vs. base and fine-tuned (FT) DiffuseMix with CLOC on IDRID.}
\label{table:dismix_vs_diffusemix_overhead}
\end{table}

\end{document}